\documentclass[10pt,twocolumn,letterpaper]{article}

\usepackage{wacv}              

\usepackage{graphicx}
\usepackage{amsmath}
\usepackage{amssymb}
\usepackage{booktabs}
\usepackage[accsupp]{axessibility}
\usepackage[table]{xcolor}

\usepackage[pagebackref,breaklinks,colorlinks]{hyperref}

\usepackage[capitalize]{cleveref}
\crefname{section}{Sec.}{Secs.}
\Crefname{section}{Section}{Sections}
\Crefname{table}{Table}{Tables}
\crefname{table}{Tab.}{Tabs.}

\DeclareMathOperator*{\argmax}{arg\,max}

\definecolor{Gray}{gray}{0.9}

\def\wacvPaperID{1858} 
\def\confName{WACV}
\def\confYear{2024}

\begin{document}

\title{Zero-shot Building Attribute Extraction from \\
Large-Scale Vision and Language Models}

\author{
\hspace{0mm}Fei Pan$^{1}$
\hspace{5mm}Sangryul Jeon$^{1}$
\hspace{5mm}Brian Wang$^{1}$
\hspace{5mm}Frank Mckenna$^{2}$
\hspace{5mm}Stella X. Yu$^{1,2}$
\vspace{1mm}
\\
$^{1}$University of Michigan, Ann Arbor
\hspace{5mm}
$^{2}$University of California, Berkeley
\\
\hspace{0mm}{\tt\small 
\{feipan, srjeon, bswang, stellayu\}@umich.edu,  fmckenna@berkeley.edu}
\\
}

\maketitle


\begin{abstract}
 Existing building recognition methods, exemplified by BRAILS, utilize supervised learning to extract information from satellite and street-view images for classification and segmentation.  However, each task module requires human-annotated data, hindering the scalability and robustness to regional variations and annotation imbalances. In response, we propose a new zero-shot workflow for building attribute extraction that utilizes large-scale vision and language models to mitigate reliance on external annotations. The proposed workflow contains two key components: image-level captioning and segment-level captioning for the building images based on the vocabularies pertinent to structural and civil engineering. These two components generate descriptive captions by computing feature representations of the image and the vocabularies, and facilitating a semantic match between the visual and textual representations. Consequently, our framework offers a promising avenue to enhance AI-driven captioning for building attribute extraction in the structural and civil engineering domains, ultimately reducing reliance on human annotations while bolstering performance and adaptability.
\end{abstract}

\section{Introduction}
\label{sec:intro}
\begin{figure}
    \centering
    \includegraphics[width=0.48\textwidth]{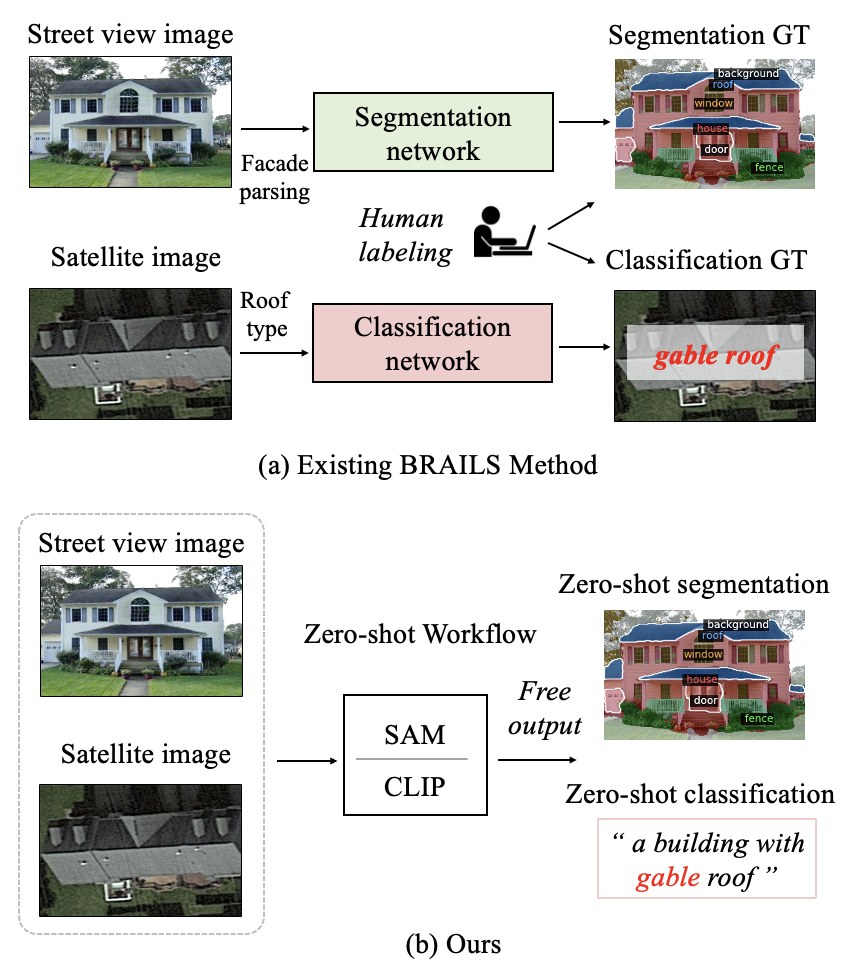}
    \caption{The existing method \textit{vs. }ours for building attribute extraction. (a) The existing prominent method BRAILS obtains building attributes from the satellite and street view images via classification and segmentation modules which require human annotation data. (b) We propose a new zero-shot workflow to extract building attributes  based on large-scale models. Our workflow uses a single module to directly extract building attributes for different tasks without human annotations and shows more robustness to novel domains.}
    \label{fig:teaser2}
\end{figure}

As a consequence of global warming, many natural hazards, such as earthquakes, hurricanes, floods, and typhoons, increase in intensity and have a destructive impact on many residential areas. To understand the impact and prepare for the potential damage caused by such hazards, the researchers in the field of natural hazards engineering and local and state agencies gather information about the buildings and other infrastructures in the areas to be studied~\cite{deierlein2019state}. Serving as a pivotal undertaking for building inventory generation and management, the task of building attribute extraction strives to provide information containing the number of floors, roof types, year of construction, etc.

Previous works~\cite{adaln,softstory} have utilized deep learning models to obtain building information from satellite and street view images acquired from mapping agencies.~\cite{yu2020rapid,yu2019building} propose to learn a model capable of identifying seismically vulnerable buildings from the street view images. These works have been collected into a software package, known as BRAILS (Building Recognition using AI at Large Scale) \cite{brails}, which extracts building attributes such as roof shape,  building height, and foundation type from satellite and street view images by employing supervised learning in the vision domain. BRAILS provides an interface allowing building inventories to be automatically generated for a region. Moreover, it contains multiple modules and each module is formulated as either image classification, object detection, or semantic segmentation. An example of BRAILS is shown in Fig.~\ref{fig:teaser2}a.

Nonetheless, the modules in BRAILS require data manually annotated by humans, particularly for the segmentation task which requires pixel-level annotations. These annotations could be obtained from different data providers or agencies, and detailed image-level descriptions are most often collected manually through a crowd-sourcing website. Nevertheless, there are several challenges that hinder the scalability and robustness of BRAILS. 
\begin{enumerate}
\setlength{\itemsep}{0pt}
\setlength{\leftmargin}{0pt}
\item The amount of human-generated annotations is inadequate to account for the large regional variations in visual and geometrical appearances, and these annotations are also prone to subjective errors. 
\item Models trained on the available data struggle to effectively generalize to novel buildings in unseen regions. 
\item The distribution of annotations across known classes may be biased or skewed, resulting in severe imbalance between minority classes of handful instances and majority classes of hundreds of instances, 
presenting a hard imbalanced classification problem.
\end{enumerate}
All these factors lead to poor model generalization performance across regions.

We propose a zero-shot workflow (Fig.~\ref{fig:teaser2}b) that tackles these challenges by utilizing large-scale models, CLIP \cite{CLIP} and SAM \cite{SAM}, which are trained on extensive and diverse datasets and can be readily adapted to downstream tasks. 
\begin{itemize}
\setlength{\itemsep}{0pt}
\setlength{\leftmargin}{0pt}
\item{CLIP} is a large-scale model that associates images with text through contrastive learning. It has the capacity to generate captions for novel images. 
\item{SAM} is a large-scale image segmentation model trained on many high-resolution images along with their annotated segmentation masks. It can provide high-quality segmentation boundaries in novel images. 
\end{itemize}

\begin{figure}
    \centering
    \includegraphics[width=0.45\textwidth]{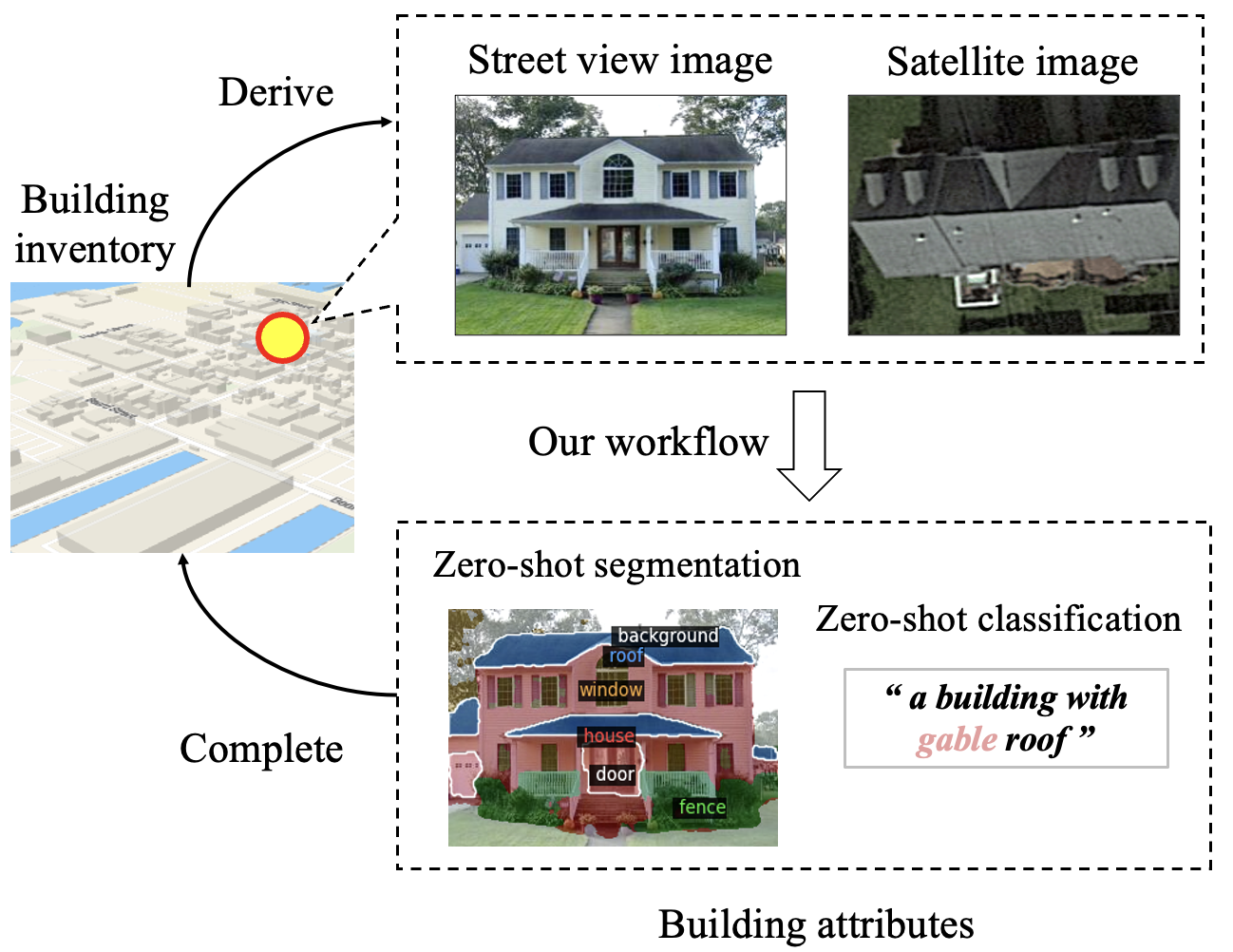}
    \caption{Our workflow directly infers building attributes on street view and satellite images. These building attributes can be used to complete the building inventory.}
    \label{fig:teaser}
\end{figure}

Our zero-shot workflow integrates the strengths of CLIP and SAM, extracting attributes of interest to structure and civil engineers without relying on human annotations. These building attributes extracted from our workflow can be used to complete the building inventory as shown in Fig.~\ref{fig:teaser}.

Our workflow has two components: image-level captioning designed for image classification, and segment-level captioning for image segmentation. 
\begin{itemize}
\setlength{\itemsep}{0pt}
\setlength{\leftmargin}{0pt}
\item In the image-level captioning component, captions are generated for building images using a list of text terms of interest to structural and civil engineers. CLIP computes feature vectors from both the image and the text terms and selects the term most similar to the image. 
\item The segment-level captioning component operates in a similar manner but begins by sending the building image to SAM first in order to obtain image segments. Then all these image segments are fed into CLIP to obtain a proper segment-level captioning.
\end{itemize}

Our work makes three major contributions. {\bf 1)} We are the first to utilize large-scale vision and language models for building attribute extraction in the structural and civil engineering domains. {\bf 2)} Our workflow facilitates zero-shot image classification and segmentation, applicable to any vocabulary of interest for building description, without reliance on human annotations. {\bf 3)} Our workflow harnesses the generalization capacity of CLIP for image captioning on building images, thereby enabling a robust and versatile building attribute extraction.




\section{Related Work}
\label{sec:related}
\begin{figure*}[t!]
    \centering
    \includegraphics[width=0.7\textwidth]{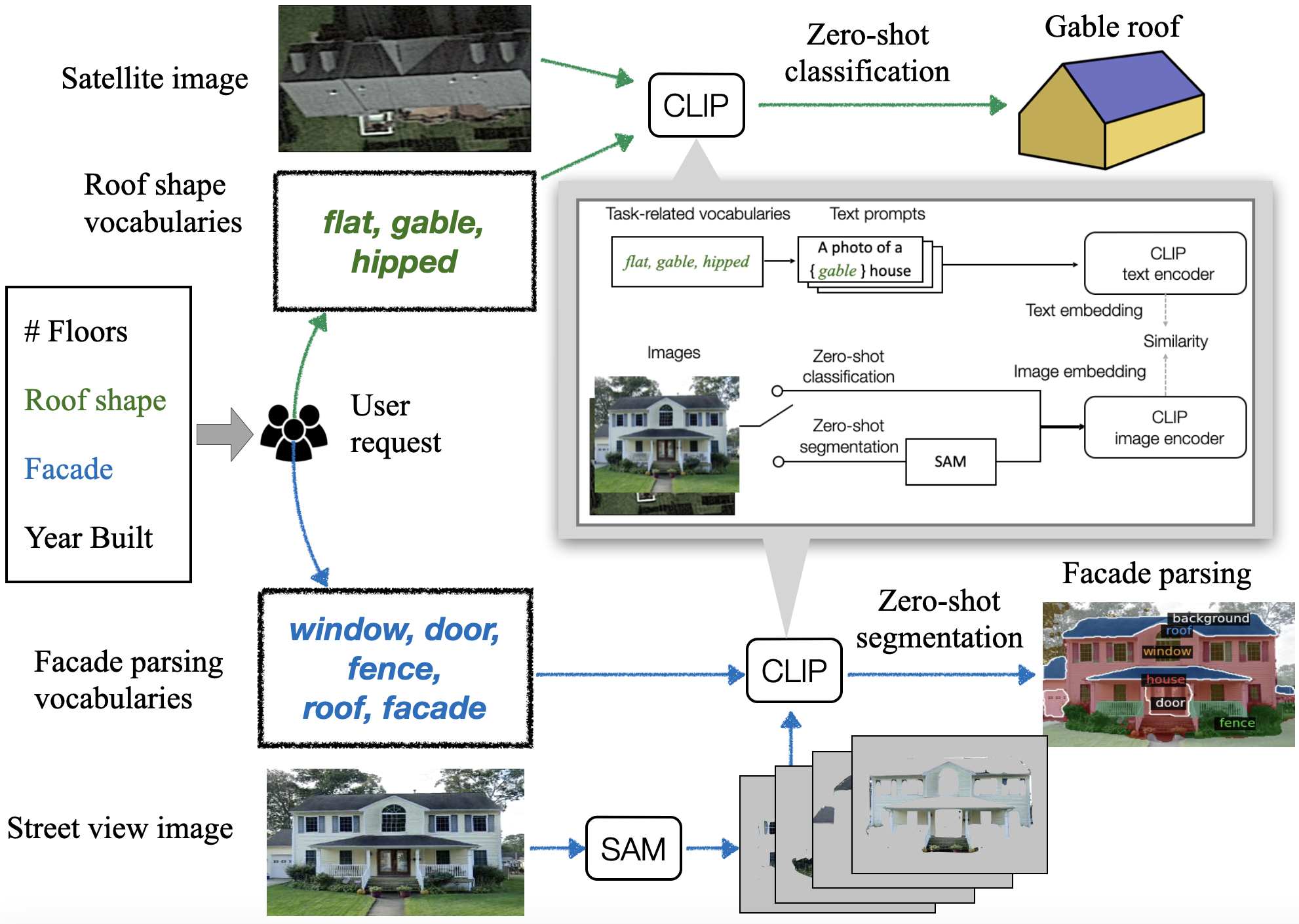}
    \caption{Our zero-shot workflow, leveraging large-scale vision and language models CLIP and SAM, effectively extracts building attributes for different tasks using the same module without human-annotated data, In contrast, traditional methods require annotated data to train multiple modules for different tasks. Our method contains two components: image-level captioning for image classification, and segment-level captioning for image segmentation for different tasks. Given a task requested by users, we first build a curated list of task-related vocabularies about building attributes. For the task processed by image-level captioning, CLIP generates a vocabulary from an image as input. For the task processed by segment-level captioning, CLIP predicts a vocabulary for each image segment produced by SAM. Benefiting from large-scale models, the proposed workflow shows strong generalization to the novel domain.}
    \label{fig:workflow}
\end{figure*}

\noindent \textbf{Learning-based building recognition.} Building information modeling~\cite{bim} in structural and civil engineering has been employed to manage buildings and infrastructures, as well as to minimize the impact of natural hazards like hurricanes, floods, and tornadoes~\cite{deierlein2019state}. Prior studies~\cite{adaln,softstory} have employed deep learning models to grasp building information from satellite and street view images sourced from map agencies.~\cite{yu2020rapid} and~\cite{yu2019building} introduce a model that learns to identify seismically vulnerable buildings from street view images.  ~\cite{gebru2017using} leverages visual cues like cars in street view images to estimate neighborhood demographics. BRAILS~\cite{brails} extracts building information from satellite and street view images using supervised learning. Each module in BRAILS is
formulated as either an image classification, object detection, or semantic segmentation task. Nonetheless, the modules within BRAILS require human-annotated data, hampering the scalability and generalization of the framework.  

Our approach tackles these challenges by making use of large-scale models already trained extensively on diverse data.  It extracts building attributes without the need for external annotations.


\noindent \textbf{Transferable features from pre-training.} Feature presentations derived from models that were pre-trained on the ImageNet~\cite{imagenet} dataset has been widely used ~\cite{maskrcnn,fasterrcnn,fcn,deeplab}. These representations have demonstrated the ability to generalize well~\cite{yosinski2014transferable} across various tasks such as object detection~\cite{maskrcnn,fasterrcnn,retinanet} and semantic segmentation~\cite{fcn,deeplab,pspnet}. Alternatively, self-supervised learning methods which involve some pretext tasks~\cite{pretext1,pretext2,pretext3}, contrastive learning~\cite{contrastive1,contrastive2,contrastive3}, clustering~\cite{clustering1}, or bootstrapping~\cite{bootstrap1}, have been employed to generate versatile feature representations. Some approaches involve learning visual representation through natural language~\cite{visuall1,visuall2,visuall3,visuall4}, where pairs of images and corresponding captions are utilized. Recent advances such as CLIP~\cite{CLIP} and ALIGN~\cite{align} have employed contrastive learning on extensive curated sets of image-text pairs, showcasing the pre-trained feature representations with remarkable zero-shot transfer capability. Specifically, the pre-trained CLIP model has been successfully used in style manipulation~\cite{styleclip}, semantic segmentation~\cite{maskclip}, panoptic segmentation~\cite{xu2023open}, and image captioning~\cite{clipscore}. 

We propose to extend the use of CLIP to the field of structural and civil engineering. By harnessing the transferable feature representations from CLIP, our approach achieves robust and accurate building attribute extraction across various residential areas and regions.

\noindent \textbf{Zero-shot learning in visual tasks.}  Most deep learning models are confined to concepts learned during training. Zero-shot learning~\cite{zeroshot_6} aims to discern novel concepts from data never seen during 
training.  ~\cite{zeroshot_1} introduces a framework for learning image and text embeddings in a joint space.  ~\cite{zeroshot_2} converts a text embedding into semantic features which are taken as guidance for segmentation.~\cite{zeroshot_3, zeroshot_4, zeroshot_5} apply a generative model to produce the semantic features of unseen classes.  

We tap into the broad visual and text knowledge CLIP and SAM have to robustly extract building attribute descriptions from images without any further training.


\section{Zero-Shot Building Attribute Extraction}
\label{sec:methodology}
Our zero-shot workflow for building attribute extraction (Fig.~\ref{fig:workflow}) 
uses pre-trained large-scale vision and language models, CLIP and SAM, to achieve robust and versatile image understanding performance without needing any human annotations. 
It has two components: image-level captioning developed for image classification, and segment-level captioning for image segmentation.

\subsection{Large-scale Models: CLIP and SAM}
As a large-scale vision and language model pre-trained on extensive sets of image-text pairs, CLIP~\cite{CLIP} comprises an image encoder denoted as $\Psi$ and a text encoder denoted as $\Phi$. These components are trained together to map input images and texts into a common representation space. The training objective of CLIP is centered on contrastive learning, taking the correct image-text pairs as positive samples while treating the mismatched pairs as negative samples.

SAM \cite{SAM} is a large-scale vision model pre-trained with more than 1 billion masks on 11 million images for image segmentation. This pre-training provides SAM with a deep understanding of various objects and scenes and strong zero-shot generalization. SAM supports multiple types of prompts such as points, boxes, and texts, and is capable of segmenting any object in an image given with certain prompts.


Given an image or an image region segmented by SAM, our method queries CLIP with a list of captions that pertain to a task in the fields of structural and civil engineering and then selects the most matching caption. This strategy allows us to extract various building attributes with a single model and strong robustness to image variations, whereas BRAILS would need one model for one task and suffer a performance drop when test images do not look like training images.

\subsection{Our Zero-shot Workflow}
\label{sec:method}
Our initial step involves obtaining the relevant vocabularies based on user-provided requests. Suppose we possess a collection of tasks pertinent to structural and civil engineering areas. Users can choose specific tasks, such as \textit{the roof shape of the building} or \textit{the semantic parsing of the facade}. Upon receiving the selected tasks, we evaluate their suitability for employment in the zero-shot classification task or the zero-shot segmentation task.
Following this, we apply the proposed zero-shot workflow to derive the building attributes associated with these chosen tasks. \\

\noindent \textbf{Zero-shot classification.}
Given a street view image $\mathbf{I}\in \mathbb{R}^{H\times W\times 3}$, we input it into a pre-trained image encoder $\Psi$ to derive its image embedding $\mathbf{E} = \Psi(\mathbf{I}) \in \mathbb{R}^{1\times L}$, where $L$ is the size of the image embedding. Let $\mathcal{T}$ represent a task from the users. We also have a group of task-specific categories $\{\mathcal{V}_1,\cdots, \mathcal{V}_{N_t}\}$ relevant to structural and civil engineering, where $\mathcal{V}_i$ represents the textual vocabulary of the $i$-th category and $N_t$ is the number of categories in the task $\mathcal{T}$. Each of these categories is integrated into the prompt and sent into a pre-trained text encoder $\Phi$ to acquire the corresponding text embedding $\mathbf{F}^{V_i} = \Phi(\mathcal{V}_i) \in \mathbb{R}^{1\times L}$. Both the image encoder and text encoder are derived from the CLIP model. Then we obtain a score vector $\mathcal{S}_i$ by computing the similarity between the image embedding $\mathbf{E}$ and the text embedding $\mathbf{F}^{V_i}$ indicated by
\begin{equation}
 \mathcal{S}_i = \texttt{sim}(\mathbf{E}, \mathbf{F}^{V_i}),
 \label{equ:sim}
\end{equation}
where $\texttt{sim}(\mathbf{E}, \mathbf{F}^{V_i})$ represents the cosine similarity, computed by the dot product between $\mathit{l}_2$-normalized $\mathbf{E}$ and 
$\mathbf{F}^{V_i}$, i.e., $\texttt{sim}(\mathbf{E}, \mathbf{F}^{V_i})= {\mathbf{E}^\top \mathbf{F}^{V_i}} / {\|\mathbf{E}\| \|\mathbf{F}^{V_i}\|}$. On this basis, we identify the most suitable category index for $\mathbf{I}$, denoted by $\mathcal{P}_{img}(\mathbf{I})$, with the highest score in $\mathcal{S}$ by
\begin{equation}
    \mathcal{P}_{img}(\mathbf{I}) = \argmax_{i} \mathcal{S}_i.
\end{equation}


\noindent \textbf{Zero-shot segmentation.}
We take advantage of SAM for segmenting residential objects including \textit{roofs}, \textit{fence}, \textit{doors}, \textit{windows}, and \textit{facades}. Let $\{ \mathcal{C}_1, \cdots,\mathcal{C}_{N_c} \}$ denote the categories for the residential objects, where $\mathcal{C}_{k}$ represents the textual vocabulary of the $k$-th category and $N_c$ is the number of categories. We take a street view image $\mathbf{I}$ as input to SAM,  which in turn produces category-agnostic non-overlapped binary masks $\mathcal{M} = \{ \mathbf{M}_j \mid \mathbf{M}_j\in\mathbb{B}^{H\times W} \}^{N}_{j=1}$, where $N$ indicates the number of the masks. Next, we need to assign these binary masks with the corresponding semantic categories. Specifically, we design a zero-shot semantic segmentation workflow using the strong image captioning capacity of CLIP. Given a binary mask $\mathbf{M}_j$, a masked image $\mathbf{I}^{M_j}$ is obtained by element-wise multiplication presented by
\begin{equation}
    \mathbf{I}^{M_j} = \mathbf{M}_j \odot \mathbf{I}.
\end{equation}
Subsequently, $\mathbf{I}^{M_j}$ is fed into the image encoder $\Psi$ to produce the masked image embedding $\mathbf{E}^{M_j} = \Psi(\mathbf{I}^{M_j})$. Simultaneously, each of the categories is incorporated into the prompt and sent into the pre-trained text encoder $\Phi$ to get the embedding $\mathbf{F}^{C_k} = \Phi(\mathcal{C}_k)$. With the masked image embedding $\mathbf{E}^{M_j}$ and the category embedding $\mathbf{F}^{C_k}$, a similarity score $\mathcal{R}_{j}$ is computed by measuring the similarity between $\mathbf{E}^{M_j}$ and $ \mathbf{F}^{C_k}$ using
\begin{equation}
    \mathcal{R}_{j,k} = \texttt{sim}(\mathbf{E}^{M_j}, \mathbf{F}^{C_k}),
\end{equation}
where $\texttt{sim}$ is defined in the same way as in Eq.~\ref{equ:sim}. Our zero-shot segmentation outputs a segmentation map $\mathcal{P}_{seg}(\mathbf{I}) \in \mathbb{R}^{H\times W}$  computed by
\begin{equation}
    \mathcal{P}_{seg}(\mathbf{I})^{(h,w)} = \argmax_{k} \big\{\mathcal{R}_{j,k} \mid \mathbf{M}_{j}^{(h,w)} = 1 \big\},
\end{equation}
where $\mathcal{P}_{seg}(\mathbf{I})^{(h,w)}$ is the predicted category index at the pixel position  $(h,w)$. Note that $\mathbf{M}_{j}^{(h,w)} = 1$ means that we take the binary mask $\mathbf{M}_{j}$ that shows $1$ at the pixel position $(h,w)$.



\section{Experiments}
\label{sec:experiments}

Our work begins by explaining how BRAILS extracts building attributes from building inventory databases. We then present a variety of tasks, including determining the number of floors, classifying roof types, identifying the year of construction, and parsing facades. BRAILS employs a random split for its training and validation datasets, evaluating on the validation set. Therefore, it presents excellent performance on the validation set but shows decline on the novel domain.  Our approach, on the other hand, uses a single module to extract building attributes for various tasks without the need for human-labeled data. We collect our own data as a novel domain and demonstrate that our method generalizes better to new domains.

\begin{figure*}
    \centering
    \includegraphics[width=0.9\textwidth]{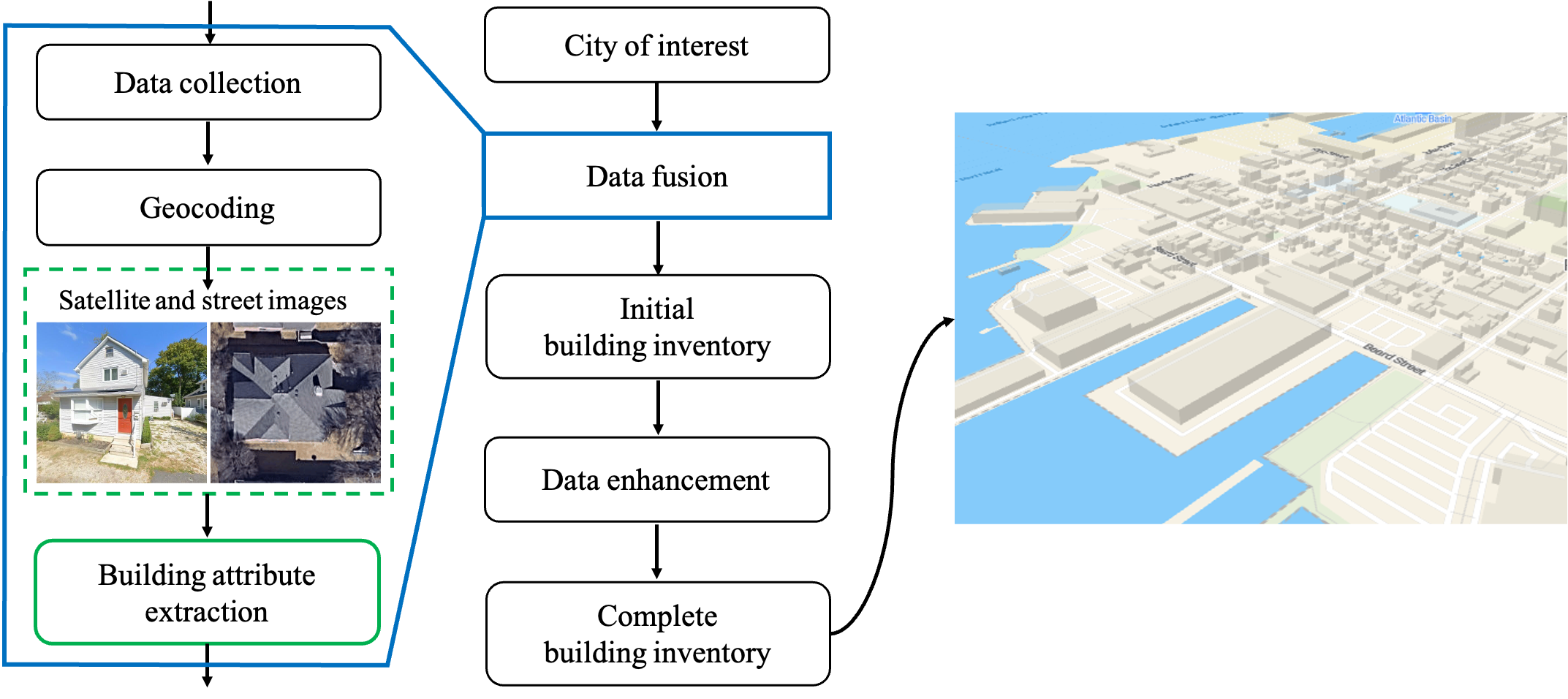}
    \caption{The framework pipeline of BRAILS offers a standardized approach to construct realistic databases of building inventories.}
    \label{fig:pipeline}
\end{figure*}

\subsection{BRAILS Framework}
BRAILS~\cite{brails2,brails3} offers a standardized approach to constructing realistic databases of building inventories, facilitating the creation of building information models at a regional scale. 
The framework of BRAILS is composed of multiple stages illustrated in Fig.~\ref{fig:pipeline}.

Collecting regional-scale building information often requires the utilization of multiple resources. These resources encompass images, point clouds, property tax records, crowd-sourced maps, etc. These resources may be owned by different entities and stored in diverse formats.  The data fusion process 
in Fig.~\ref{fig:pipeline} 
aims to create fused building information data that surpasses the original data in terms of informativeness and comprehensiveness.

The data collection module in the data fusion involves the integration of multiple building information datasets to yield information that is more consistent, precise, and practical than what any single source can provide. The outcome of data collection still lacks building attribute information due to data scarcity. For instance, crowd-sourced maps often lack completeness, particularly in rural areas, while they tend to be more comprehensive in densely urbanized regions and areas targeted for humanitarian mapping interventions. Similarly, administrative databases containing property tax assessment records often exhibit missing entries. This deficiency is a common issue across nearly all data sources. To handle these issues, the BRAILS framework applies the building attribute extraction module (Sec.~\ref{sec:data_att_ext}) based on the initial database to predict the absent building attributes and effectively fill in the missing values.

The initial building inventory generated by the data fusion process might still remain incomplete.  For example, occlusions caused by trees or cars can impact predictions related to building attributes, such as the number of floors. This occlusion can subsequently lower the precision of predictions obtained from the building attribute extraction model. Therefore, a data enhancement module is implemented to address these incomplete predictions, aiming to make valuable contributions toward achieving a comprehensive final building inventory.

\subsection{Building Attribute Extraction}
\label{sec:data_att_ext}
We assume the preliminary collection of indexing information including building addresses, coordinates, year of construction, and structural style was established. 
With the indexing information in hand, it becomes feasible to utilize the Google Maps API for retrieving satellite and street-view images corresponding to each respective building.

The existing BRAILS framework comprises multiple modules to extract building attributes (absent in the original database) from these images. These modules leverage deep learning architectures and are designed for the image classification task and the semantic segmentation task, refined through supervised learning.

\begin{figure}[t!]
    \centering
    \includegraphics[width=0.45\textwidth]{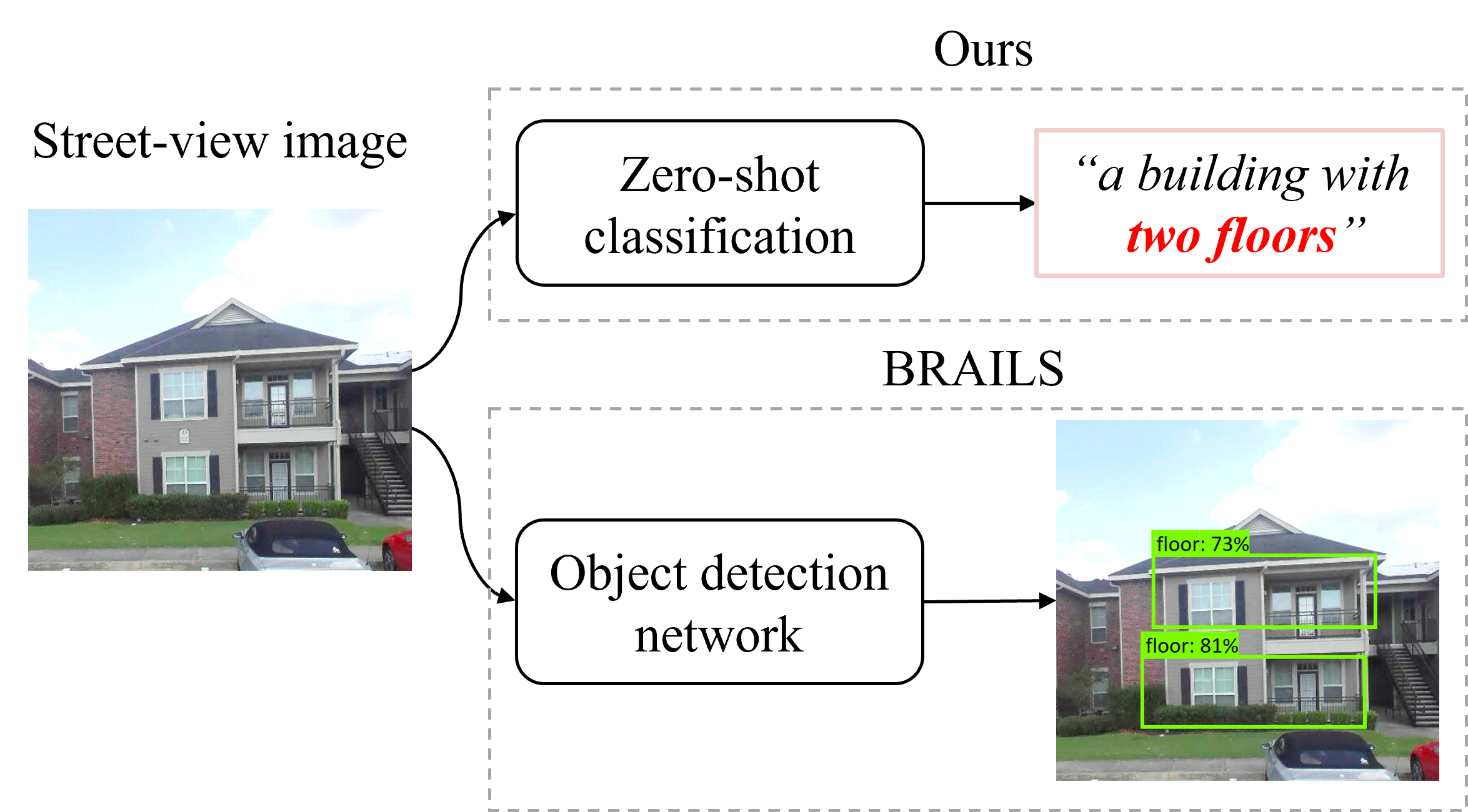}
    \caption{BRAILS requires a supervised object detection network to detect the number of floors, whereas our zero-shot workflow is capable of making direct inferences on the input images without any fine-tuning.}
    \label{fig:num_of_floors}
\end{figure}

\begin{table}[t!]   
    \caption{BRAILS presents superior scores on the roof type classification task. Note that the BRAILS's scores are essentially training accuracy due to a lack of split over the training and the validation set. Our score is approaching BRAILS on the flat type. }
    \label{tab:roof_type_classification}
    \centering
    \resizebox{0.40\textwidth}{!}{
    \begin{tabular}{r | r r r| r}
        \hline
        Accuracy ($\%$) & \# Images & BRAILS & Ours & \textit{Our Gain}\\
        \hline
        Gable type & 8449 & \textbf{99.2} & 2.0 & -97.2 \\
        Hip type & 8451 & \textbf{99.4} & 47.8 & -51.6 \\
        Flat type & 8447 & \textbf{99.6}  & 98.1 &  -1.5 \\
        \hline
        \rowcolor{Gray} Micro-Average &  & \textbf{99.4}  & {49.2} & -50.2 \\
        \rowcolor{Gray} Macro-Average & & \textbf{99.4}  & {49.2} & -50.2 \\
        \hline
    \end{tabular}
    }
\end{table}

However, the BRAILS method for building attribute extraction has two main limitations. First, it requires human annotation for all tasks, and these annotations can be expensive and time-consuming. Second, the performance of the models in BRAILS is evaluated on validation datasets that are similar to the training set. However, these models' generalization is not carefully studied. The ability of these models to generalize to new and unseen data from different geographical locations depends on how similar the building inventory for these locations is to the data used to train the BRAILS models. Generalization is a complex topic that has been actively researched recently. In this work, we propose a new zero-shot workflow that addresses these challenges by using large-scale models that are trained with self-supervised techniques on a variety of datasets and can be adapted to a wide range of downstream tasks.


\subsection{Roof Type Classification} 
\begin{figure}[t!]
    \centering    \includegraphics[width=0.45\textwidth]{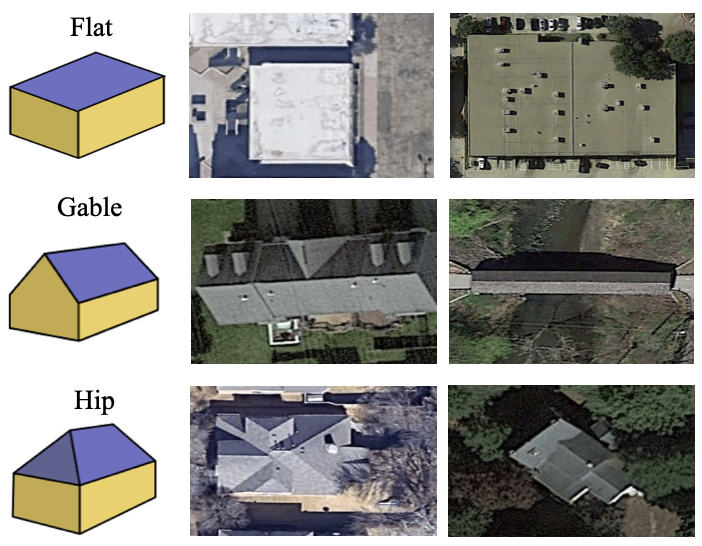}
    \caption{The common roof types and example satellite images for these types.}
    \label{fig:roof_type}
\end{figure}

\noindent \textbf{Dataset.} We present the three primary categories of roof types commonly used globally: flat, gabled, and hipped (Fig.~\ref{fig:roof_type}). The whole dataset collected in BRAILS contains $6,000$ labeled satellite images with $2,000$ images for each of the three roof types. The dataset is randomly split into the training and validation set. 

\noindent \textbf{Baseline.} To identify the roof types of every building in the region, BRAILS adopts an \textit{image classification network} with a ResNet-50~\cite{resnet}  backbone. Different from the roof type classification module in BRAILS, our workflow treats this task as a \textit{zero-shot classification} task and does not require pre-training on these training data.



\noindent \textbf{Results.}

Table~\ref{tab:roof_type_classification} shows results on different roof types: gable, hip, and flat types. Note that due to the lack of train-test split information, the results from BRAILS in Table~\ref{tab:roof_type_classification} represent the training accuracy. On average for roof type classification, BRAILS presents $99.4\%$ of accuracy whereas our method indicates a performance of $49.2\%$. Our score is approaching BRAILS on the flat type. There are two reasons for this performance gap. First, BRAILS utilizes a supervised image classification model and encounters a small domain gap between the training and validation set. In contrast, our method is a zero-shot workflow that does not require any pre-training on annotated data. Second, our workflow utilizes the image captioning ability of CLIP which is less generalized to satellite images.

\begin{table}[t!]   
    \caption{The scores of BRAILS and ours for the year built classification are computed using BRAILS' own validation set. BRAILS presents superior performance because it adopts supervised training with numerous human-annotated data, whereas our method requires neither human annotations nor any fine-tuning.}
    \label{tab:year_built_classification}
    \centering
    \resizebox{0.415\textwidth}{!}{
    \begin{tabular}{r | r r r | r}
        \hline
        Year range & \# Images & BRAILS & Ours & \textit{Our Gain}\\
        \hline
        Pre 1969 & 30198 & \textbf{62.0} & 38.7 & -23.3 \\
        1970 - 1979 & 10485 & \textbf{11.6} & 0.8 & -10.8 \\
        1980 - 1989 & 20519 & 10.8 & \textbf{12.8} &  +2.0\\
        1990 - 1999 & 13537 & 8.3  & \textbf{46.3} &  +38.0\\
        2000 - 2009 & 19178 & \textbf{14.0}  & 0.1 &  -13.9\\
        Post 2010 & 5944 & \textbf{1.6}  & 0.0 &  -1.6\\
        \hline
        \rowcolor{Gray} Micro-Average &  & \textbf{26.1}  & 20.7 & -5.4 \\
        \rowcolor{Gray} Macro-Average &  & \textbf{18.1}  & 16.4 & -1.7 \\
        \hline
    \end{tabular}
    }
\end{table}

\begin{figure*}
    \centering
    \includegraphics[width=.8\textwidth]{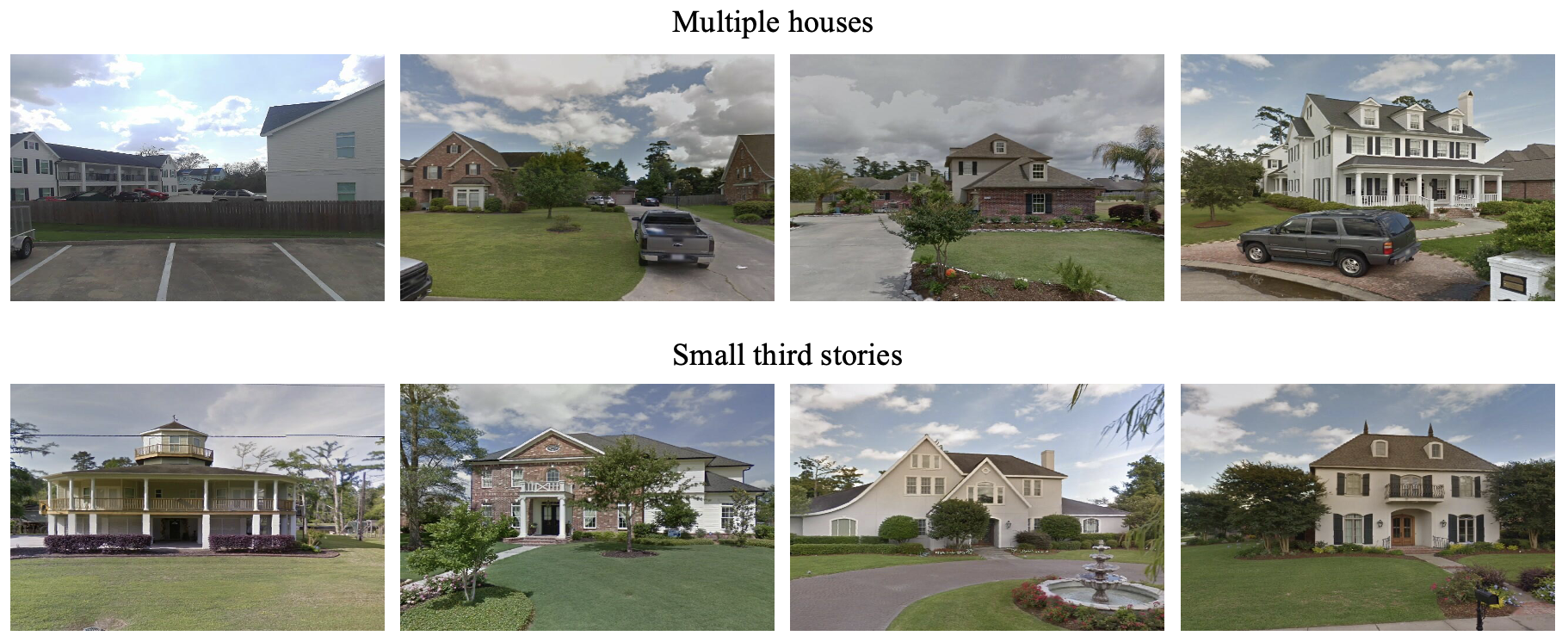}
    \caption{The exemplary images of the three-story houses from BRAILS' validation set where our method completely failed.}
    \label{fig:example_images}
\end{figure*}

\subsection{Year Built Classification} 
\noindent \textbf{Dataset.} The objective of the year built classification task involves categorizing buildings into various groups, each representing a distinct range of construction years. The dataset devised for year built classification consists of $56,660$ annotated street-view images and has a random split for training and validation set. We consider 6 distinct ranges of year built constructions:  before 1969 (Pre 1970), 1970-1979, 1980-1989, 1990-1999, 2000-2009, and after 2010 (Post 2010). We compare ours against BRAILS on the BRAILS' own validation set for the year built classification task. 

\noindent \textbf{Baseline.}
In the BRAILS framework, this task is considered by using an \textit{image classification network} with ResNet-50~\cite{resnet} as the backbone. In contrast, our approach adopts a \textit{zero-shot classification} method, omitting any need for supervised learning with the training data.


\noindent \textbf{Results.} 
In Table~\ref{tab:year_built_classification}, on average for the year built classification, the BRAILS method presents an accuracy of $26.1\%$, while our method shows $20.7\%$. This performance gap reflects that the BRAILS is trained in such annotated data with a random training and testing split and is not affected by the domain gap between the training and validation set. \textit{Our proposed zero-shot workflow requiring neither annotations nor training already gets close to BRAILS' supervised learning method on the year-built classification task.}


\subsection{Facade Parsing}
\noindent \textbf{Dataset \& Task.} Building facade parsing needs to segment the building facade from a street view image into multiple semantic categories, such as the roof, windows, doors, and facades. This can be used to extract more detailed building attributes and contribute to a complete building inventory. The existing BRAILS method for facade parsing requires pixel-wise annotations from humans, which can be costly and time-consuming. 
\begin{figure}[t!]
    \centering
    \includegraphics[width=.48\textwidth]{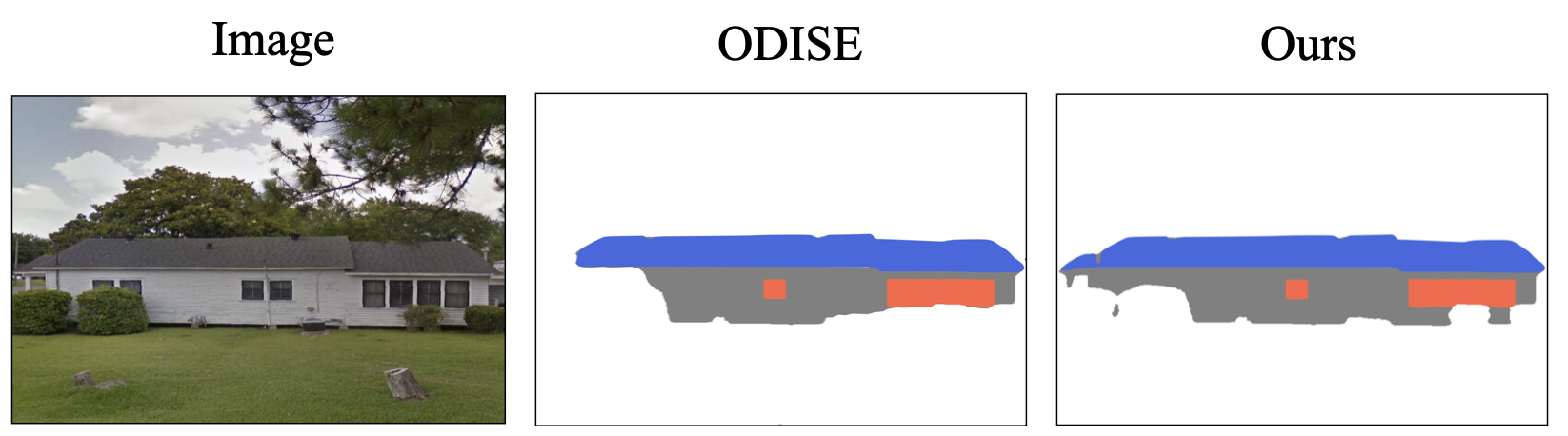}
    \caption{The qualitative results of our model against ODISE~\cite{odise} for facade parsing.}
    \label{fig:fp_vis}
\end{figure}

\begin{table}[t]   
    \caption{Our method is compared with existing open-vocabulary segmentation methods OVSeg and ODISE on facade parsing task. The mIoU ($\%$) performance is evaluated on the validation set of BRAILS.}
    \label{tab:facade_parsing}
    \centering
    \resizebox{0.415\textwidth}{!}{
    \begin{tabular}{r | r r r r | r}
        \hline
        mIoU ($\%$) & Roof & Door & Window & Facade & \textit{Mean} \\
        \hline
        OVSeg~\cite{ovseg} & 50.9 & 66.4 & 79.1 & 38.8 & 57.3 \\
        ODISE~\cite{odise} & 53.1 & 64.1 & 83.0 & 44.7 & 60.2 \\
        \hline
        \rowcolor{Gray} Ours & \textbf{55.6} & \textbf{67.9} & \textbf{85.5} & \textbf{48.6} & \textbf{61.5} \\
        \hline
    \end{tabular}
    }
\end{table}

\noindent \textbf{Baseline.} We provide an ablation study of our zero-shot segmentation on the dataset in BRAILS for the facade parsing task. We choose recent state-of-the-art open-vocabulary semantic segmentation methods OVSeg~\cite{ovseg} and ODISE~\cite{odise} as our baseline models. OVSeg requires large amounts of annotated data and requires fine-tuning over CLIP model, and ODISE involves a training guided by the text-to-image diffusion UNet~\cite{diffusion_unet}.

\noindent \textbf{Results.} Table~\ref{tab:facade_parsing} shows the comparison results of ours with the baseline models OVSeg and ODISE. Our method presents a mIoU (mean intersection over union) of $61.5$ which is the highest among all the baseline methods that require annotated data or extra guidance. This is due to the strong image segment capability from SAM model in our zero-shot segmentation. \textit{It indicates that we apply zero-shot segmentation to achieve facade parsing without requiring annotated data or additional fine-tuning. Moreover, it allows us to segment images that have never been seen before, even if the building images are from different styles or regions.} The qualitative results of our method against ODISE are shown in Fig.~\ref{fig:fp_vis}.



\subsection{\# Floors}
\begin{figure}
    \centering
    \includegraphics[width=0.48\textwidth]{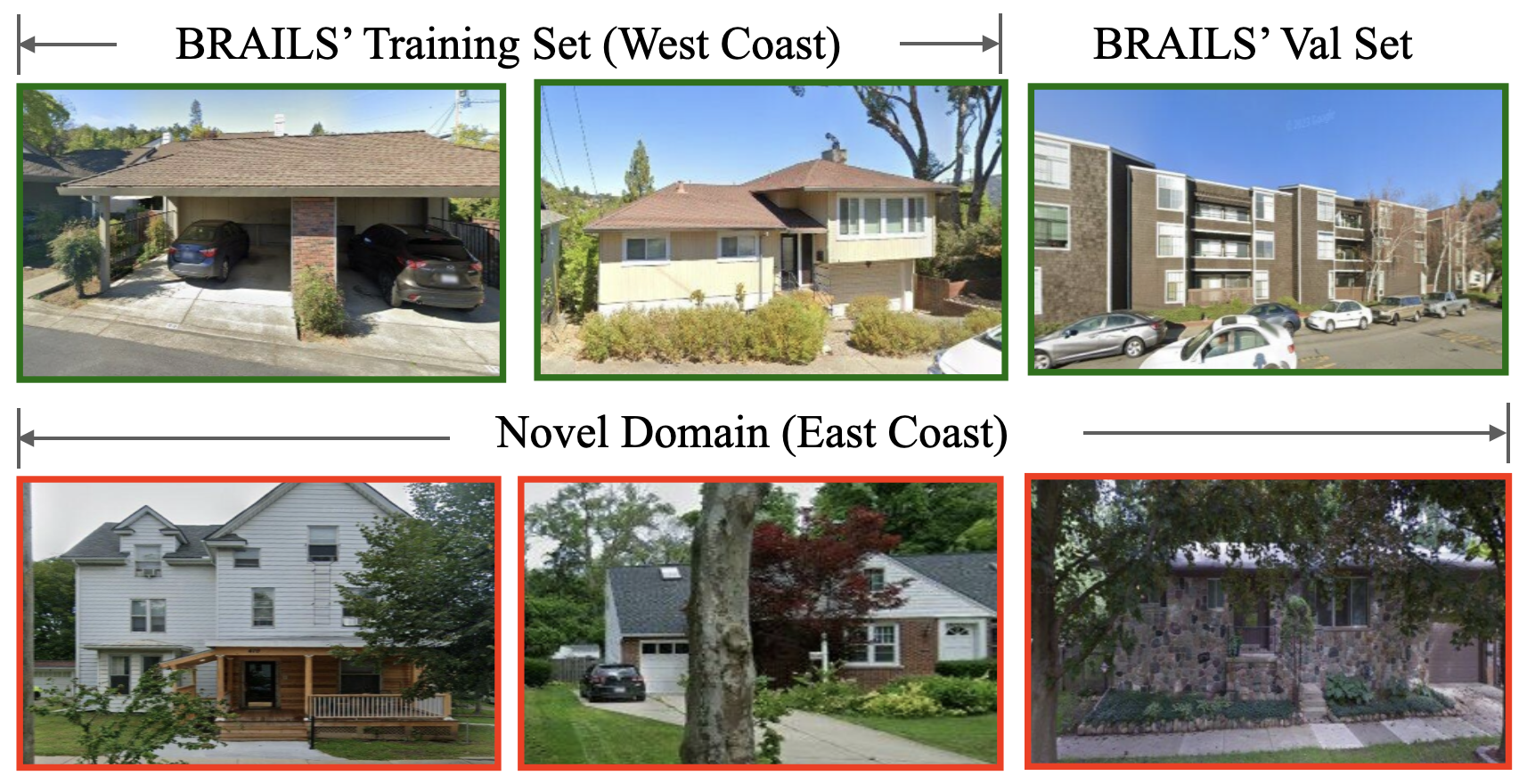}
    \caption{The first-row images are from BRAILS' dataset collected at West Coast regions, while the second-row images are from our novel domain collected in East Coast regions. BRAILS performs \textcolor{green}{well} (\textcolor{red}{poorly}) when the test data (\textcolor{red}{does’t}) \textcolor{green}{matches} the training data.}
    \label{fig:num_floors}
\end{figure}

\label{sec:tasks}
\begin{table}[t!]   
    \caption{The scores of BRAILS and ours for detecting the number of floors are evaluated on BRAILS' own validation set. BRAILS shows superior performance because it adopts supervised training and encounters few domain gaps with the validation set, whereas our method requires neither human annotation nor additional fine-tuning.}
    \label{tab:num_floor_classification}
    \centering
    \resizebox{0.415\textwidth}{!}{
    \begin{tabular}{r | r r r | r}
        \hline
        Accuracy ($\%$) & \# Images & BRAILS & Ours & \textit{Our Gain}\\
        \hline
        One-story & 2393 & \textbf{88.5} & 80.8 & -7.7  \\
        Two-story & 580 & 56.4 & \textbf{57.8} & +1.4 \\
        Three-story & 16 & \textbf{56.3}  & 0.0 &  -56.3\\
        \hline
        \rowcolor{Gray} Micro-Average &  & \textbf{82.0}  & {75.9} & - 6.1 \\
        \rowcolor{Gray} Macro-Average &  & \textbf{67.0}  & {46.2} &  -20.8 \\
        \hline
    \end{tabular}
    }
\end{table}

\vspace{-4pt}

\begin{table}[t!]   
    \caption{ BRAILS suffers from a clear performance drop due to the domain gap from the novel domain, while our method presents a robust performance on the novel domain data for detecting the number of floors. }
    \label{tab:roof_type_novel_domain}
    \centering
    \resizebox{0.42\textwidth}{!}{
    \begin{tabular}{r | r r r | r}
        \hline
        Accuracy ($\%$) & \# Images & BRAILS & Ours & \textit{Our Gain}\\
        \hline
        One-story & 210 & 70.7 & \textbf{77.6} & +6.9\\
        Two-story & 198 & 55.2 & \textbf{74.0} & +18.8 \\
        Three-story & 37 & 33.3 & \textbf{50.0} & +16.7\\
        \hline
        \rowcolor{Gray} Micro-Average & & 66.5 & \textbf{76.1} & +9.6  \\
        \rowcolor{Gray} Macro-Average & &  53.07 & \textbf{67.2} & +14.13  \\
        \hline
    \end{tabular}

    }
\end{table}

\noindent \textbf{Dataset.} The dataset collected in BRAILS comprises $60,000$ street view images sourced from various counties in New Jersey in the United States, excluding Atlantic County. These data are then partitioned into three subsets: a training set, a validation set, and a testing set, randomly distributed in proportions of $80\%$, $15\%$, and $5\%$ of the total data, respectively. 

\noindent \textbf{Baseline.} Within the BRAILS framework, an ~\textit{object detection network} based on EfficientDet-D4 architecture~\cite{efficientdet} is employed to identify visible floors in the street-view images. In contrast, our workflow tackles this task as a \textit{zero-shot classification} problem and we eliminate the need for any preliminary training on the training set as shown in Fig.~\ref{fig:num_of_floors}.

\noindent \textbf{Results.} According to Table~\ref{tab:num_floor_classification}, on BRAILS' own validation set, the average accuracy of BRAILS is $82.0\%$ while the accuracy of our method is $75.9\%$. The difference in accuracy between the two methods is due to the fact that the BRAILS method is a supervised object detection model trained on multiple data with annotated bounding boxes. In contrast, our method is a zero-shot workflow that does not require any training or human annotations. We completely failed on three-story houses in Table~\ref{tab:num_floor_classification}. The reasons are threefold: 1) there are only a few images for the three-story house; 2) some images contain multiple houses; 3) the third stories are usually very small. We present the exemplary images for these failing cases in Fig.~\ref{fig:example_images}.

%
\noindent \textbf{Generalization.} We further evaluate the generalization of the BRAILS method and ours over novel domains. As BRAILS takes data that mostly comes from the cities on the West Coast of the U.S. for training, it is reasonable that BRAILS shows good performance in these regions. We postulate that there exists domain gap between West Coast and East Coast images. Therefore, we select some image data from the cities on the East Coast as a novel domain. Specifically, we collect street view images from Houston, Texas and randomly select 460 images as the novel domain for testing. We provide some sample images from BRAILS' dataset and our novel domain in Fig.~\ref{fig:num_floors}The results of the BRAILS method and our method on the novel domain are shown in Table~\ref{tab:roof_type_novel_domain}. Our method presents an average accuracy of $76.1\%$ and BRAILS shows $66.5\%$ on the data from the novel domain. This result confirms that BRAILS suffers from a generalization gap due to regional variations in building appearances. \textit{Our proposed zero-shot workflow provides a more robust, promising baseline to novel domains without model supervision or prompt tuning.} \\

\section{Conclusion}
\label{sec:conclusion}
In this paper, we proposed a new zero-shot workflow for building attribute extraction in the structural and civil engineering domains. Our workflow utilizes large-scale vision and language models, CLIP and SAM, to generate captions for buildings from satellite and street-view images. This allows us to extract building attributes without relying on human annotations. Our workflow has several advantages over existing methods. First, it is scalable and robust to regional variations in visual and geometrical appearances. Second, it can generalize to novel buildings in unseen regions. We evaluated our workflow on the datasets of building images with the task of zero-shot image classification and zero-shot segmentation. Our results demonstrate the effectiveness of our approach for building attribute extraction. In the future, we plan to take a more advanced way of utilizing large-scale models and turn our zero-shot workflow into a specific expert model in structural and civil engineering areas.

{\noindent\bf Acknowledgements.}
This work is funded in part by NSF 2131111: Natural Hazards Engineering Research Infrastructure: Computational Modeling and Simulation Center.

{\small
\bibliographystyle{ieee_fullname}
\bibliography{egbib}
}

\end{document}